\documentclass[graybox]{svmult}
\usepackage[first=0,last=9]{lcg}
\usepackage{scalerel}   

\usepackage{wrapfig}

\usepackage{mathptmx}       
\usepackage{helvet}         
\usepackage{courier}        
\usepackage{makeidx}         
\usepackage{graphicx}        
\usepackage{multicol}        
\usepackage{wrapfig}
\usepackage{amstext} 
\usepackage{array}   
\newcolumntype{C}{>{$}c<{$}} 

\makeindex             

\usepackage{amssymb}
\usepackage{amsmath}
\usepackage{algorithm}
\usepackage{algorithmic}

\DeclareMathOperator*{\argmin}{argmin} 
\usepackage{multirow}
\usepackage{array}
\usepackage{tikz}           
\usepackage{color, colortbl}
\usepackage{makecell}
\usepackage{todonotes}

\definecolor{Gray}{gray}{0.9}

\begin{document}

\title*{Machine Learning in LiDAR 3D point clouds}

\author{F. Patricia Medina, Randy Paffenroth}

\institute{
F. Patricia  Medina \at Yeshiva University; \email{patricia.medina@yu.edu} \\
Randy Paffenroth \at Worcester Polytechnic Institute; \email{rcpaffenroth@wpi.edu}  
}

\maketitle


\abstract{LiDAR point clouds contain measurements of complicated natural scenes and can be used to update digital elevation models, glacial monitoring, detecting faults and measuring uplift detecting, forest inventory, detect shoreline and beach volume changes, landslide risk analysis, habitat mapping and urban development, among others. A very important application is the classification of the 3D cloud into elementary classes. For example, it can be used to differentiate between vegetation, man-made structures and water. Our goal is to present a preliminary comparison study for classification of 3D point cloud LiDAR data that includes several types of feature engineering.  In particular, we demonstrate that providing context by augmenting each point in the LiDAR point cloud with information about its neighboring points can improve the performance of downstream learning algorithms.  We also experiment with several dimension reduction strategies, ranging from Principal Component Analysis (PCA) to neural network based auto-encoders, and demonstrate how they affect classification performance in LiDAR point clouds. For instance, we observe that combining feature engineering with a dimension reduction method such as PCA, there is an improvement in the accuracy of the classification with respect to doing a straightforward classification with the raw data.}

\section{Introduction}

LiDAR point clouds contain measurements of complicated natural scenes and can be used to update digital elevation models, glacial monitoring, detecting faults and measuring uplift detecting, forest inventory, detect shoreline and beach volume changes, landslide risk analysis, habitat mapping and urban development, among others. A very important application is the classification of the 3D cloud into elementary classes. For example, it can be used to differentiate between vegetation, man-made structures and water.

This paper describes results from using several classification frameworks in 3D LiDAR point clouds. We present a preliminary comparison study for classification of 3D point cloud LiDAR data. We experiment with several types of feature engineering by augmenting each point in the LiDAR point cloud with information about its neighboring points and also with dimension reduction strategies, ranging from Principal Component Analysis (PCA) to neural network based auto-encoders, and demonstrate how they affect classification performance in LiDAR point clouds. We present $F_1$ scores for each of the experiments, accuracy and error rates to exhibits the improvement in classification performance. Two of our proposed frameworks showed a big improvement in error rates. 

LiDAR is an active optical sensor that transmits laser beams towards a target while moving through specific survey routes. The reflection of the laser from the target is detected and analyzed by receivers in the LiDAR sensor. These receivers record the precise time from when the laser pulse leaving the system to when it returns to calculate the range distance between the sensor and the target, combined with the positional information GPS (Global Positioning System), and INS (inertial navigation system). These distance measurements are transformed to measurements of actual three-dimensional points of the reflective target in object space. See \cite{inbook} and \cite{Mather2004} for a technical treatment of remote sensing.

Deep learning for 3D point clouds has received a lot of attention due to its applicability to various domains such as computer vision, autonomous driving and robotics. The most common tasks performed are  3D shape classification \cite{shape}, 3D object detection and tracking \cite{tracking}, and 3D point cloud segmentation \cite{segmentation}.  Key challenges in this domain include the high dimensionality and the unstructured nature of 3D point clouds. In the case of 3D shape classification, recent methods include: projection-based networks (multi-view representation and volumetric representation) \cite{Volumetric01,Volumetric02} and point--based networks (point-wise MLP networks,convolution-based networks, graph-based networks and others) \cite{graphbased}. See \cite{unknown} for a comprehensive survey in deep learning for 3D point clouds. This paper describes results from different classification frameworks in 3D LiDAR point clouds in relevant classes of a natural scene. Note that our goal is to classify point by point instead of performing shape classification and we develop a preliminary framework to gain understating of the performance of specific combinations of algorithms applied to a specific LiDAR point cloud dataset.

Our framework includes engineering new features from existent ones, possible non-linear dimensionality reduction (auto-encoders), linear dimensionality reduction (PCA) and finally the use of a feed-forward neural network classifier.  The outputs of these preprocessing steps are then used as training data for a number of classifications algorithms including random forest and k-nearest neighbor classifiers. 

\begin{figure}
\includegraphics[scale=0.5]{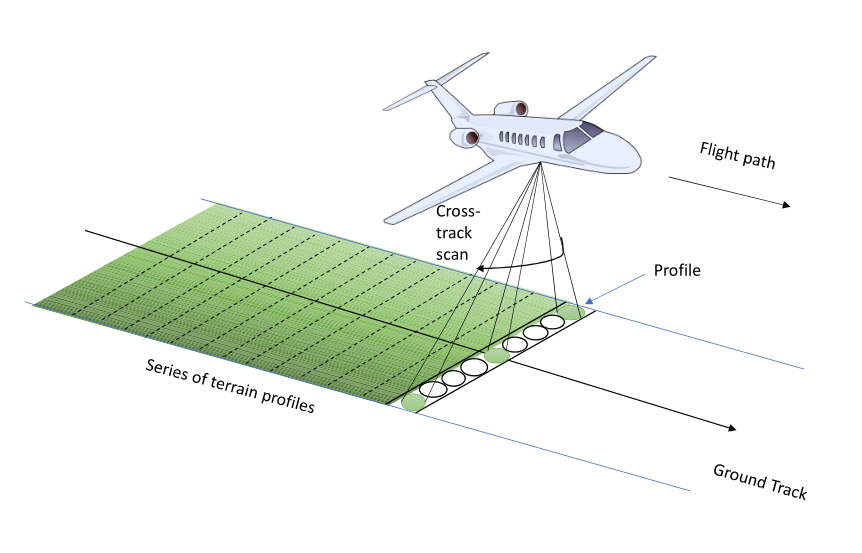}
\caption{The profile belonging to a series of terrain profiles is measured in the cross track direction of an airborne platform.  The image was recreated from figure 1.5 (b), pp. 8 in \cite{Mather2004}. The figure was used for the first time in one of the authors' paper (see \cite{medina2019heuristic}.)}
\label{fig: airplane}
\end{figure}

LiDAR stands for light detection and ranging and it is an optical remote sensing technique that uses laser light to densely sample the surface of the earth, producing highly accurate $x,\,y$ and $z$ measurements. The resulting mass point cloud data sets can be managed, visualized, analyzed and shared using ArcGIS \cite{arcgis}. The collection vehicle of LiDAR data might be an aircraft, helicopter, vehicle or tripod. (See Fig.\ref{fig: airplane})

LiDAR can be applied, for instance, to update digital elevation models, glacier monitoring, detecting faults and measuring uplift detecting, the forest inventory, shoreline detection, measuring beach volume changes, landslide risk analysis, habitat mapping and urban development \cite{Mather2004,inbook}.

3D LiDAR point clouds have many applications in the Geosciences. A very important application is the classification of the 3D cloud into elementary classes. For example, it can be used to differentiate between vegetation, man-made structures and water. Alternatively, only two classes such as ground and non-ground could be used. Another useful classification is based on the heterogeneity of surfaces. For instance, we might be interested classifying the point cloud of reservoir into classes such as gravel, sand and rock. The design of algorithms for classification of this data using a multi-scale intrinsic dimensionality approach is of great interest to different scientific communities. See the work in \cite{BRODU} and \cite{BaIzMcNeSh:2012} for classification of a natural scene using support vector machines. We also refer the interested reading to \cite{medina2019heuristic} which multi-scale testing of a multi-manifold hypothesis where LiDAR data is used as a case study and intrinsic dimension is computed.


The paper is organized as follows. First, in section \ref{sec: the data} the
attributes of LiDAR data are described. In section \ref{sec: the data}, we
provide the formal classification code for each class in Table~\ref{table: classes}	. In section \ref{sec: neigh matrix} we describe the construction of the neighbor matrix, which is a
way of generating a new data frame using the original features of the nearest
neighbors of each design point. Next, in section \ref{sec: frameworks}, we
briefly describe the machine learning frameworks used in our experiments and
define the metric uses in our experiments.  Three of the frameworks the
construction of a neighbor matrix as a way of feature engineering. Two of the
latter frameworks include linear dimension reduction (PCA) or non-linear
dimension reduction (auto-encoder.) In Section \ref{sec: experiments}, we
describe the experiments, give a more detailed description of each
classification framework, and  provide a summary of the $F_1$ scores in Table
\ref{main table}. Section \ref{summary} summarizes the results and proposes some future research directions.

\section{The data}\label{sec: the data}

LiDAR points can be classified into a number of categories including bare earth or ground, top of canopy, and water (see Fig.\ref{fig: classes}). The different classes are defined using numeric integer codes in the LAS files. Classification codes were defined by the American Society for Photogrammetry and Remote Sensing (ASPRS) for LAS formats. In the most update version eighteen classes were defined and it includes

\begin{table}
	\begin{center}		
		\begin{tabular}{c|r l}
			0   &Never classified & \\
			1   & Unassigned & \\
			2  & Ground & $\xleftarrow{\hspace*{3cm}}$\\
			3 & Low vegetation & \\
			4 & Medium vegetation & \\
			5 & High vegetation & \\
			6 & Building & \\
			7 & Noise &  $\xleftarrow{\hspace*{3cm}}$\\
			8 & Model key/ Reserved & \\
			9 & Water & $\xleftarrow{\hspace*{3cm}}$ \\
			10 & Rail &  $\xleftarrow{\hspace*{3cm}}$ \\
			11 & Road surface & \\
			$\vdots$  &  $\vdots$ & \\
			17 & Bridge deck  & $\xleftarrow{\hspace*{3cm}}$\\
			18 & High noise &  $\xleftarrow{\hspace*{3cm}}$ \\
		\end{tabular}
	\end{center}
	\caption{Classification codes. The arrows are pointed towards the six classes (ground, noise, water, rail, bridge deck and high noise we use for our experiments in the LiDAR data set graphed in Fig.~\ref{fig: long island}. See  \url{https://desktop.arcgis.com/en/arcmap/10.3/manage-data/las-dataset/lidar-point-classification.html} for a complete class code list. }
 \label{table: classes}	
\end{table}

\begin{table}
\begin{center}		
\begin{tabular}{l |r }
Class code & Number of points\\
\hline	
2   &  1578495\\
17  &  1151888\\
18  &   611201\\
9   &  226283\\
10  &   21261\\
7   &     462\\
\hline

\end{tabular}
\end{center}
\caption{Number of points used per class. See Table~\ref{table: classes} for class codes.}
\label{fig: class-points}
\end{table}

\begin{figure}
	\includegraphics[trim={0.3cm 0cm 0 0cm},clip,scale=0.6]{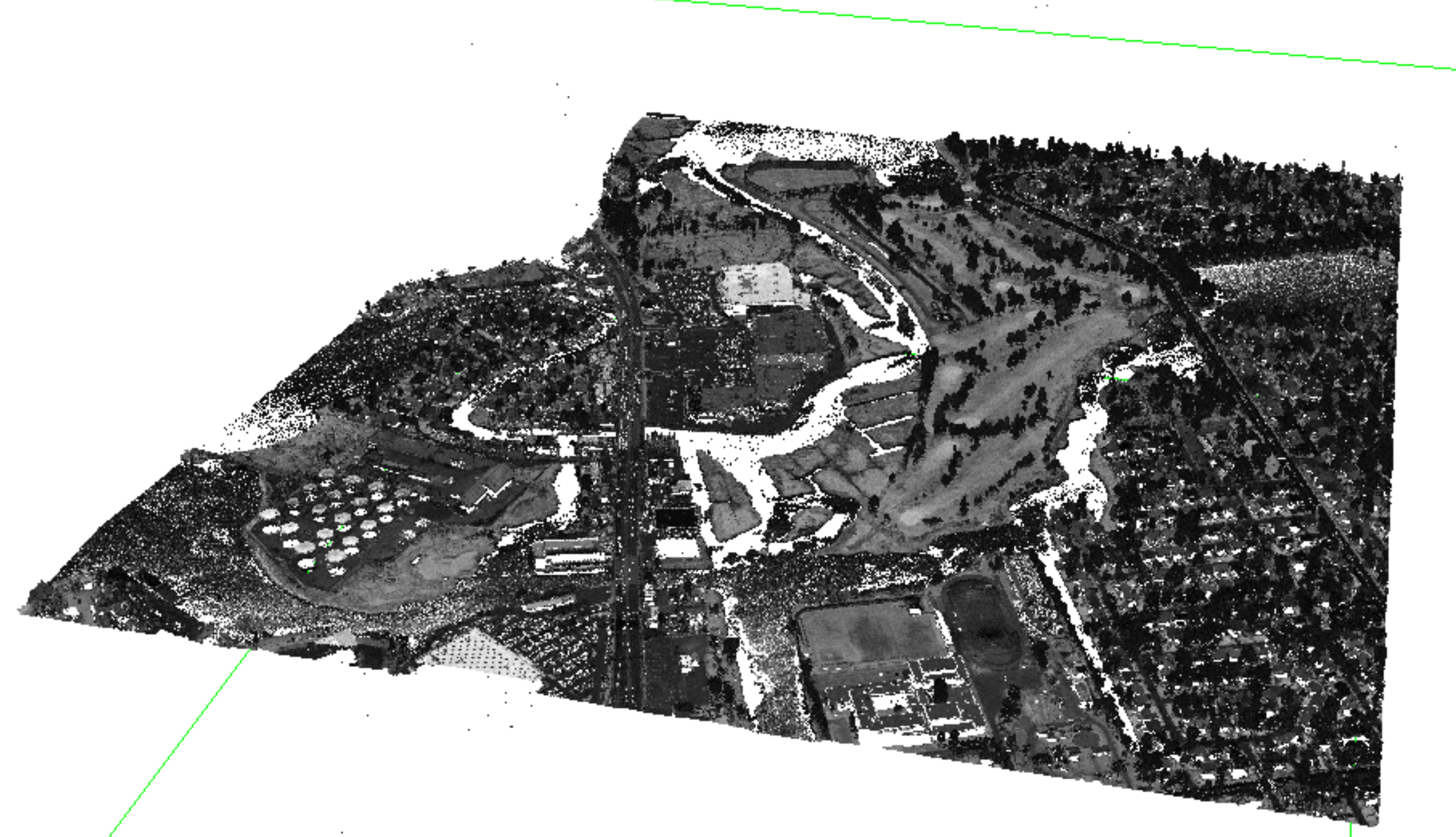}
	\caption{3D LiDAR point cloud graphed by intensity for a location close to the JFK airport, NY.}
	\label{fig: long island}
\end{figure}

In our experiments, we use a publicly available LiDAR data set (USGS Explorer)from a location close to the JFK airport. We used the visualization toll from the LAS tool \cite{LAStools} to graph the scene by intensity (see Fig. \ref{fig: long island}.) The data consists of $5.790384e \times 10^6$ points. We work with six classes (See codes in Table \ref{fig: classes}.)  The unassigned classification class is not providing any useful information for training the learning algorithm. We decided to consider the six remaining classes. Note that noise points are the ones which typically have a detrimental impact on data visualization and analysis. For example, returns from high-flying birds and scattered pulses that encountered cloud cover, smog haze, water bodies, and highly reflective mirrors can distort the z-range of the points surrounding that location.

\begin{figure}
	\includegraphics[trim={0.3cm 0cm 0 0cm},clip,scale=0.6]{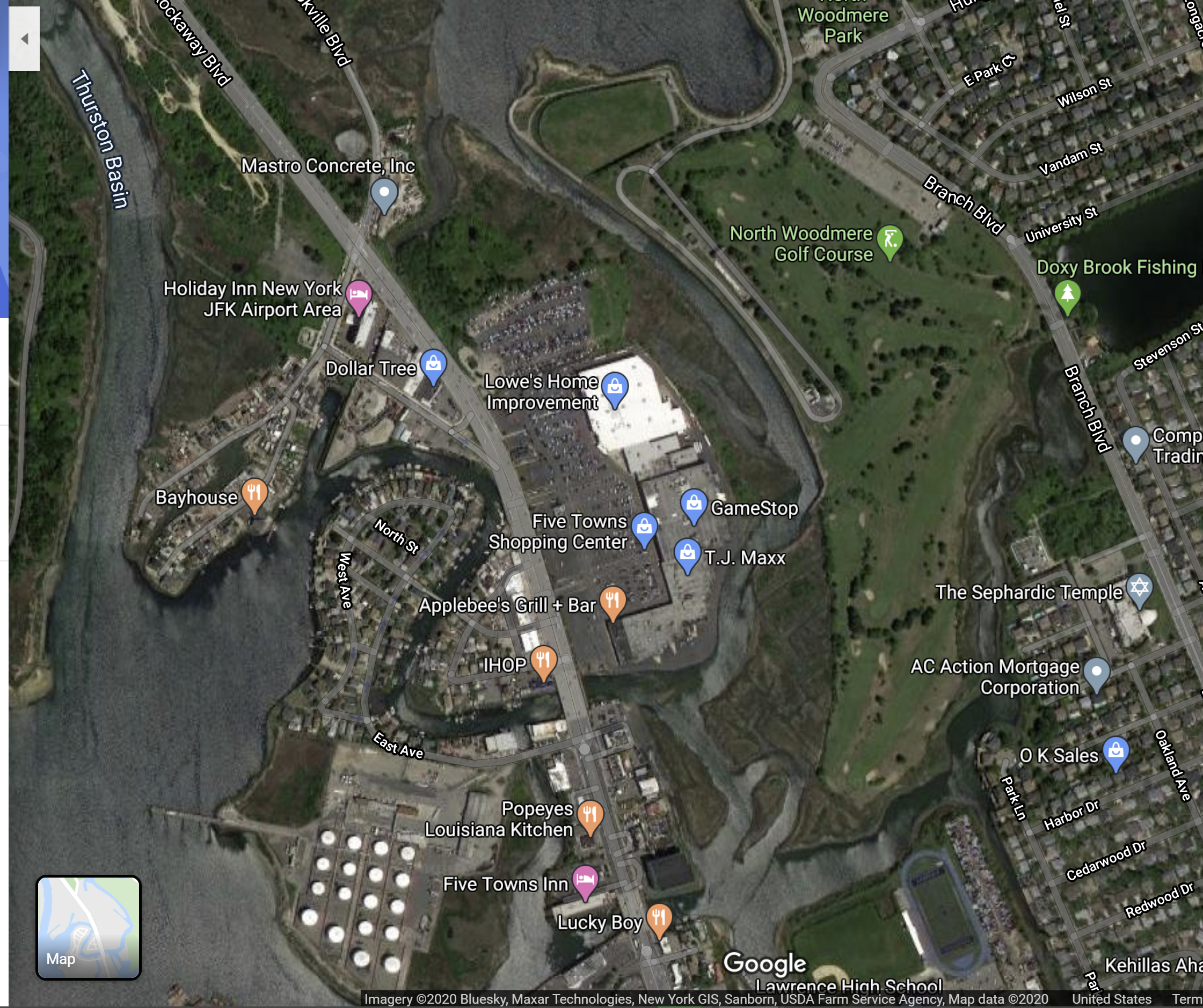}
	\caption{Google map \cite{googlemaps} satellite image of the location of associated to the 3D point cloud in the JFK airport, NY. Coordinates: $40^\circ38'38.6"N 73^\circ 44'46.9"W$
		Rockaway Blvd, Rosedale, NY 11422 See Fig.\ref{fig: long island}. Link to the exact location: \url{https://goo.gl/maps/aWa47Gxzb5wuYNu76}}
	\label{fig: google maps}
\end{figure}

We included a snapshot of the satellite view from Google maps in Fig.\ref{fig: google maps}. The geographical information in LiDAR is given in UTM. 

\begin{figure}
	\includegraphics[trim={0cm 7cm 0 0},clip,scale=0.6]{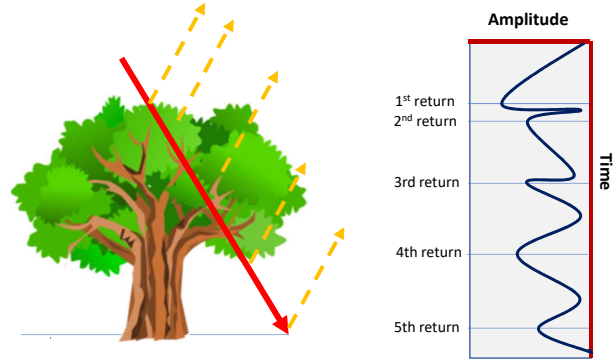}
	\caption{A pulse can be reflected off a tree’s trunk, branches and foliage as well as reflected off the ground. The image is recreated from figure from p.7 in \cite{inbook}.}
	\label{fig: tree}
\end{figure}

The following attributes along with the position $(x,y,z)$ are maintained for each recorded laser pulse. We stress that we are working with airborne LiDAR data and not terrestrial LiDAR (TLS.) 
\begin{enumerate}
	\item Intensity. Captured by the LiDAR sensors is the intensity of each return. The intensity value is a measure of the return signal strength. It measures the peak amplitude of return pulses as they are reflected back from the target to the detector of the LiDAR system. 
	\item Return number. An emitted laser pulse can have up to five returns depending on the features it is reflected from and the capabilities of the laser scanner used to collect the data. The first return will be flagged as return number one, the second as return number two, and so on. (See Fig.\ref{fig: tree}) Note that for TLS we only have one return so this attribute would not be used in that case.

	\item Number of returns. The number of returns is the total number of returns for a given pulse. Laser pulses emitted from a LiDAR system reflect from objects both on and above the ground surface: vegetation, buildings, bridges, and so on. One emitted laser pulse can return to the LiDAR sensor as one or many returns. Any emitted laser pulse that encounters multiple reflection surfaces as it travels toward the ground is split into as many returns as there are reflective surfaces. (See Fig.\ref{fig: tree})
	
	\item Point classification. Every LiDAR point that is post-processed can have a classification that defines the type of object that has reflected the laser pulse. LiDAR points can be classified into a number of categories including bare earth or ground, top of canopy, and water. The different classes are defined using numeric integer codes in the LAS files.

Airborn LiDAR data is usually collected into surface data products at local and regional level. The data is collected and post-processed by a very specialized and expensive software that is not available to the general public. One of the attributes produced in the post-processing phase is ``classification''. Many users are not able to extract directly classes from the the LiDAR point cloud due to the lack of accessibility of such commercial software. This classification is not always to be trusted and a machine learning algorithms for automated classification would simplify this task for user reduces costs. (See \cite{ramamurthy2015geometric}.)

	\item Edge of flight line. The points will be symbolized based on a value of 0 or 1. Points flagged at the edge of the flight line will be given a value of 1, and all other points will be given a value of 0.
	\item RGB. LiDAR data can be attributed with RGB (red, green, and blue) bands. This attribution often comes from imagery collected at the same time as the LiDAR survey.
	\item GPS time. The GPS time stamp at which the laser point was emitted from the aircraft. The time is in GPS seconds of the week.
	\item Scan angle. The scan angle is a value in degrees between -90 and +90. At 0 degrees, the laser pulse is directly below the aircraft at nadir. At -90 degrees, the laser pulse is to the left side of the aircraft, while at +90, the laser pulse is to the right side of the aircraft in the direction of flight. Most LiDAR systems are currently less than $\pm$30 degrees.
	\item Scan direction. The scan direction is the direction the laser scanning mirror was traveling at the time of the output laser pulse. A value of 1 is a positive scan direction, and a value of 0 is a negative scan direction. A positive value indicates the scanner is moving from the left side to the right side of the in-track flight direction, and a negative value is the opposite.
\end{enumerate}

In all of our experiments we only keep a total of seven attributes: x, y,z,intensity,scan angle,number of returns,number of this return. Note that RGB values can be obtained form satellite map images such as Google maps. We decided not perform the data integration step to include these values since we prefer to work with only the original LiDAR data set (see Fig.\ref{fig: long island}).

\section{Feature engineering: nearest neighbor matrix}\label{sec: neigh matrix}

We uniformly select s examples out of the original data. For each LiDAR data point (example) we consider $k$ nearest neighbors based on spatial coordinates $(x_i,y_i,z_i)$ and create a new example which is in higher dimensions. The new example we generated includes all the features of all neighbors (not only the spatial features.)

More precisely, let $F_{n(0)}^{(i)}$ the set of $N$ features associated to the $i$th example (the first three features are spatial.) Now let  $F_{n(j)}^{(i)}$ the set of $N$ features associated to the $j$th  nearest neighbor to the $i$th example.  So if we consider the first $k$th nearest neighbors (computed respect to the spatial features), we end up with set of set of features associated to the $i$th example:

\begin{equation}
F_{n(0)}^{(i)}, F_{n(1)}^{(i)}, \ldots , F_{n(k)}^{(i)},
\label{eqn: row neigh}
\end{equation}

where $i=1,\ldots, s.$ Here $F_{n(j)}^{(i)} \in \mathbb{R}^{1 \times N}$ for each $j=1,\ldots, k.$

We concatenate the features in \eqref{eqn: row neigh} and obtain rows 

\begin{equation}
\left[
\begin{array}{c|c|c|c}
F_{n(0)}^{(i)} &  F_{n(1)}^{(i)} &  \ldots & F_{n(k)}^{(i)} 
\end{array}
\right] 
\in \mathbb{R}^{1 \times (k+1)\cdot N}
\end{equation}

for each $i=1, \ldots, s.$ We then put all the rows together and get what we call the {\it neighbor matrix} in \eqref{eqn:neigh2}

\begin{equation}
\left[
\begin{array}{c|c|c|c}
F_{n(0)}^{(1)} &  F_{n(1)}^{(1)} &  \ldots & F_{n(k)}^{(1)}\\
F_{n(0)}^{(2)} &  F_{n(1)}^{(2)} &  \ldots & F_{n(k)}^{(2)}\\
\vdots         & \vdots          &   \vdots & \vdots\\
F_{n(0)}^{(s)} &  F_{n(1)}^{(s)} &  \ldots & F_{n(s)}^{(1)}\\
\end{array}
\right]
\in \mathbb{R}^{s \times (k+1)\cdot N}
\label{eqn:neigh2}
\end{equation}

\begin{figure}
\hspace{-1cm}\includegraphics[scale=0.5, trim={2cm 4cm 2cm 2cm},clip]{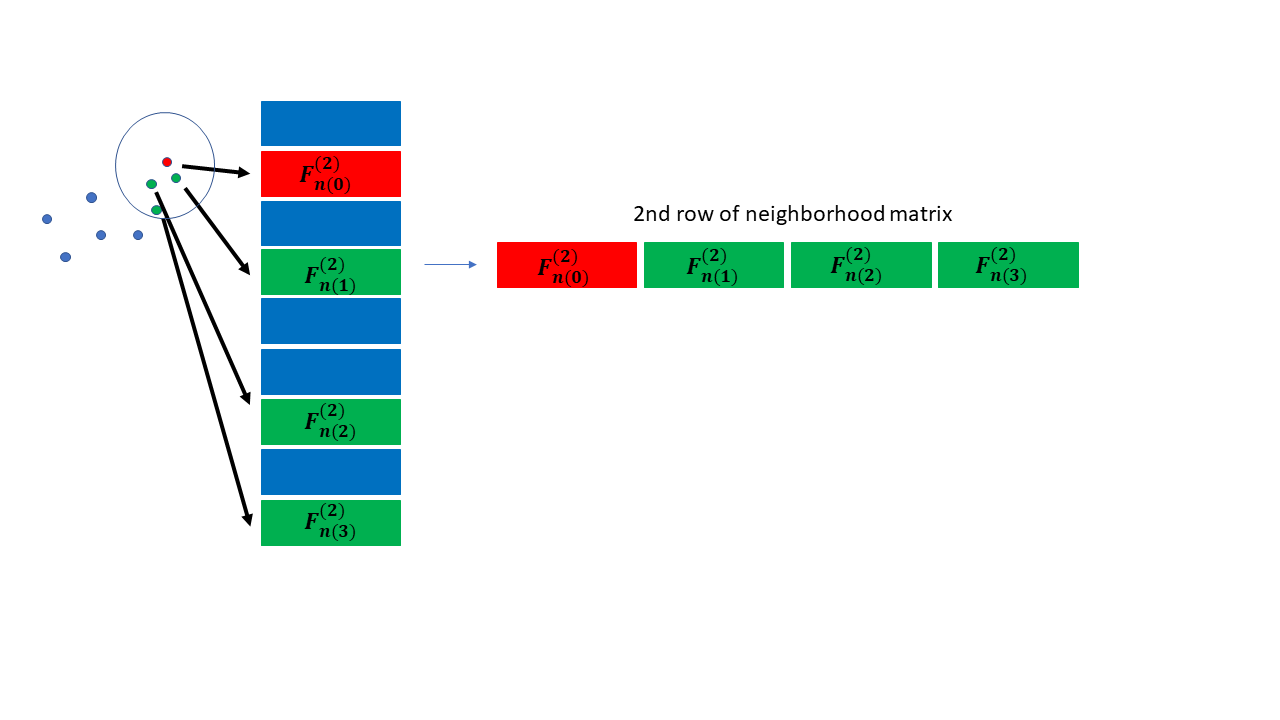}
\caption{Forming the second row by concatenating the features of of the 3 nearest neighbors to the the second example in the original data frame. The neighbors are computed respect to the spatial coordinates $(x,y,z)$ of the design point. We are working with the list of features presented in \eqref{eqn: row neigh} for $i=2$ and $k=3.$ See also the second row of the matrix in \eqref{eqn:neigh2}. Observe that if the original data has $N=7$ features, the neighbor matrix has $(3+1)\times 7=28$ features. }
\label{2nd row neigh}
\end{figure}

We illustrate how to obtain the second  row of the neighbor matrix in Fig. \ref{2nd row neigh}.

Observe that in Fig.~\ref{2nd row neigh}, $F_{n(1)}^{(2)}$ can also be a design point $F_{n(0)}^{(4)}$ and it could share nearest neighbors with the design point $F_{n(0)}^{(2)}.$
In our experiments described in section~\ref{sec: experiments}, we chose $s=100,000$ construct the neighbor matrix.

\section{Machine learning frameworks}\label{sec: frameworks}

Two of our frameworks use the neighbor matrix described in section \ref{sec: neigh matrix}   as  input. We design a machine learning algorithm for our neighbor matrix. We summarize the steps for the frameworks with dimensional reduction step. First, perform dimensionality reduction wither using either PCA (for a linear projection) or an auto-encoder. If using PCA, then use the projected features as the predictors for our learning algorithm (classifier.) If using an auto-encoder, then use the inner layer as the predictor for our classifier. Last,  provide the projected training sample (labeled) to a classifier. We use K-nearest neighbor (KNN) and Random Forrest classifiers (RF and RF-Ens), feed forward neural network (NN).


The metric that we use to measure precision of our algorithm is given by
\begin{equation}
PRE_{micro}=\dfrac{\sum_{j=1}^N TP_j}{\sum_{j=1}^N TP_j+ \sum_{j=1}^N FP_j},
\end{equation}

(known as micro average)
where $TP_i$ means true positive on the $ith$ class and $FP_i$ means false positive on the $ith$ class.

The recall (or sensitivity) is given by 
\begin{equation}
Recall=\dfrac{\sum_{j=1}^N TP_j}{\sum_{j=1}^N TP_j + \sum_{j=1}^N FN_j},
\end{equation}
where $FN_j$ means false negative on the $jth$ class.

We provide the

\begin{equation}
 F_1 \mbox{ score }= 2 \dfrac{PRE_{micro} \cdot Recall}{PRE_{micro} + Recall}, 
 \label{f1-score}
\end{equation}

Using the $F_1$-scores as metric, the learning algorithm including the auto-encoder to perform dimensionality reduction performs better than the one that feeds the classifier with the projected features resulting from performing PCA.

\begin{figure}	
\includegraphics[scale=0.7, trim={8cm 4cm 0 0},clip]{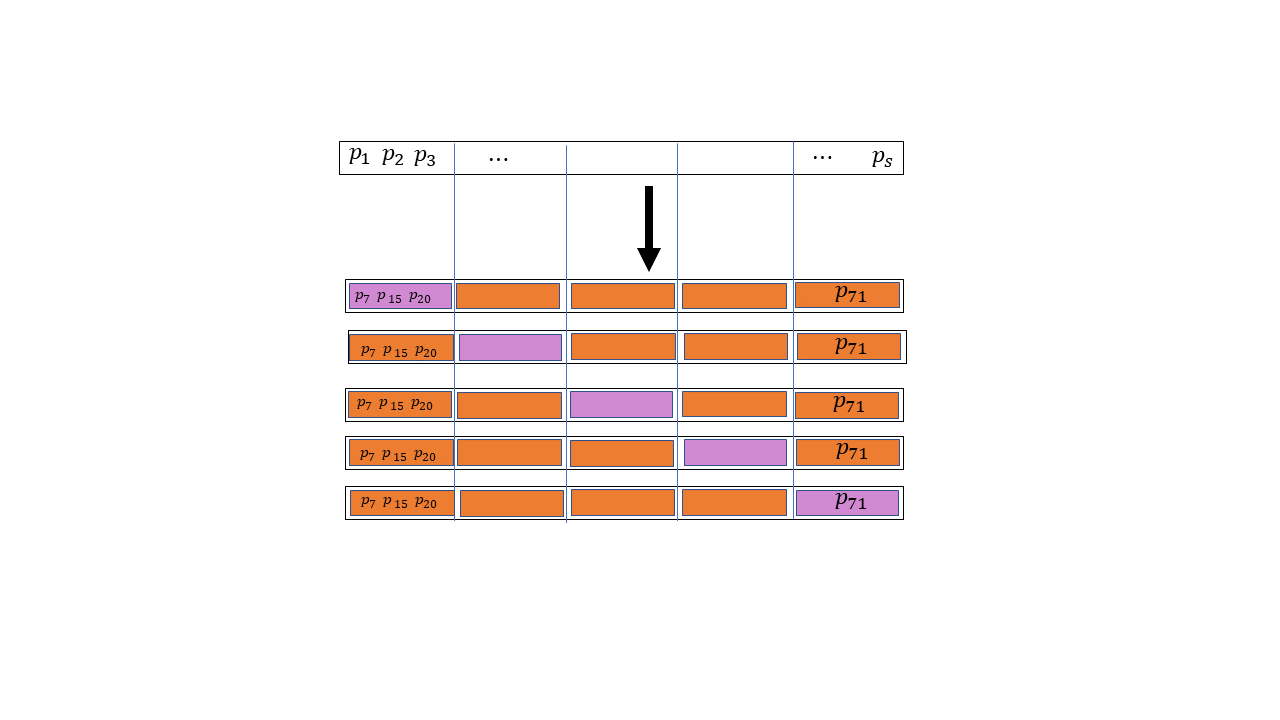}
\caption{5-fold CV example for $s$ data points: $p_1,p_2, \ldots, p_s$. Each randomly selected fifth is used as a validation set (shown in purple), and the remainder as a training set (shown in orange). The $F_1$ score is computed for each split and then the mena of the $F_1$ scores is computed. The CV scores calculated as in \ref{eqn: CV scores}. Such scores for experiments described in Section~\ref{sec: experiments} are summarized in Table\ref{main table}. The figure is a recreation of a graph from \cite{hastie_09_elements-of.statistical-learning}, p. 181.}
\label{5-fold}
\end{figure}

We use a K-fold cross validation score with the $F_1$ scores. The general idea 
is to randomly divide the data into $K$ equal-size parts. We leave out part $k$, fit the model to the other $K-1$ parts (combined), and then obtain predictions for the left-out $k$th part.  This is done in turn for each part $k=1,2,\ldots K$ , and then the results are combined. See \cite{hastie_09_elements-of.statistical-learning} for a more detailed description of re-sampling methods. Fig.~\ref{5-fold} illustrate the 5-fold re-sampling procedure. 

The scores in Table \ref{main table} are the 

\begin{equation}
\mbox{ mean of the CV score } \pm 2\times \mbox{standard deviations of the CV score},
\label{eqn: CV scores}
\end{equation}

where CV scores means the 5-fold cross validation score for $F_1$ scores.

We used TensorFlow (an open source software library for numerical computation using data flow graphs, see \cite{tensorflow}) to build the auto-encoder. The rest of the scripts are in Python using Sci-kit Learn \cite{scikit} and Pandas \cite{pandas} libraries.

In all experiments in Section~\ref{sec: experiments}, we performed the final classification stage with K-nearest neighbors (KNN), random forest (RF), and ensemble of random forest (RF-Ens) and a (layer) feed forward neural network (NN.) We standardized and normalized the input data for all of our experiments.

\subsection{Dimension reduction}\label{sec: dim reduct}

We chose PCA among the unsupervised linear methods and an auto-encoder as an unsupervised non-linear method to perform dimension reduction. Recall that we inserted a dimension reduction stage in some of our frameworks (see Section~\ref{table: experiment-description} for experiment descriptions including dimension reduction methods.)

PCA is one of the most popular unsupervised learning techniques and it performs linear dimensionality reduction that preserves as much of the variance in the data as possible after embedding the data into a linear subspace of lower dimension. The interested reader can look the detailed exposition in \cite{hastie_09_elements-of.statistical-learning}.

Deep auto-encoders are feed-forward neural networks with an odd number of hidden layers and shared weights between the left and right layers \cite{10.5555/2207825}. The input data $X$ (input layer) and the output data $\hat{X}$  (output layer) have $d^{(0)}$ nodes (the dimension of the layer.) More precisely, auto-encoders learn a non-linear map from the input to itself through a pair of encoding and decoding phases \cite{Zhou-Paffenroth2017}

\begin{equation}
\hat{X}=D(E(X)),
\end{equation}
where $E$ maps the input layer $X \in \mathbb{R}^{d(0)}$ to the ``most'' hidden layer ({\it encodes} the input data) in a non-linear fashion, $D$ is a non-linear map from the ``most'' hidden layer to the output layer (decodes the ``most'' hidden layer), and $\hat{X} $ is the recovered version of the input data. In a 5-layer auto-encoder $\hat{X} \in \mathbb{R}^{d(3)}$. An auto-encoder therefore solves the optimization problem:

\begin{equation}
\argmin_{E,\,D} {\| X - D(E(X)) \|}_2^2,
\end{equation}

We are motivated to include deep auto-encoders (or multilayer auto-encoders) in our experiments, since they demonstrated to be effective for discovering non-linear features across problem domains.

\begin{figure}
\hspace{-1.5cm}
\includegraphics[scale=0.55, trim={4cm 2cm 0cm 0cm},clip]{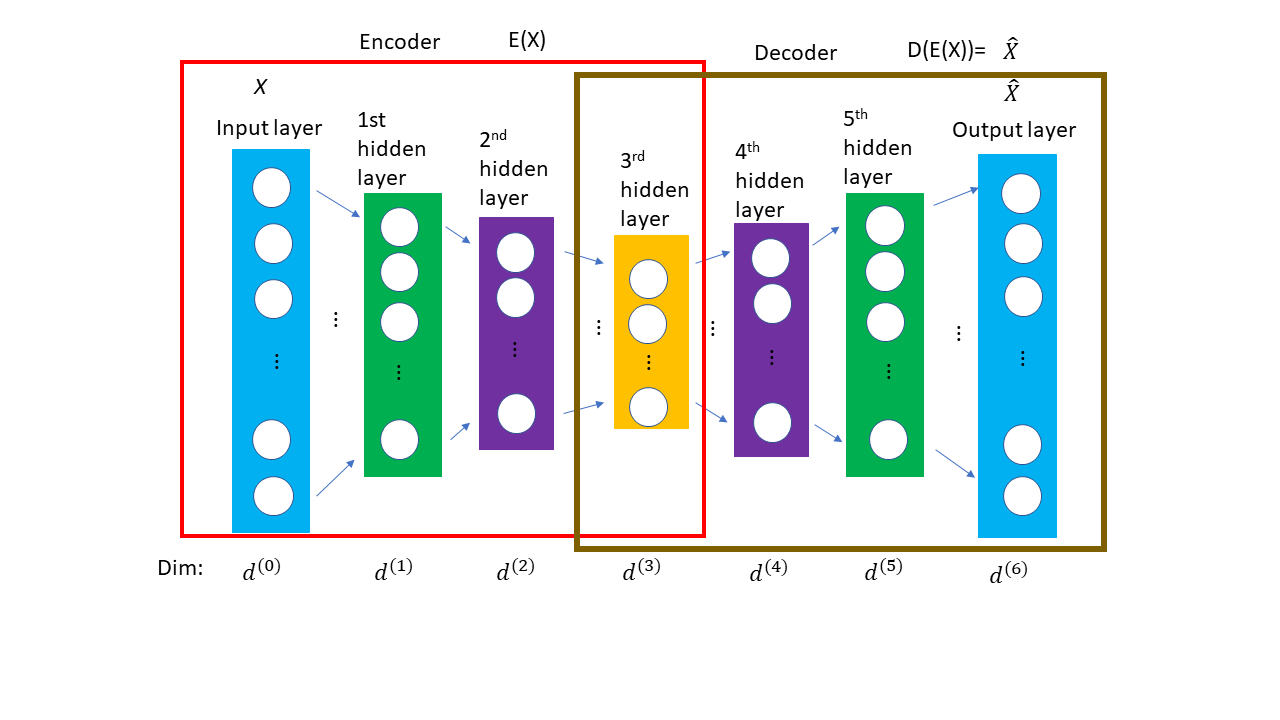}
\caption{5-layer auto--encoder diagram. The input layer has dimension $d^{(0)}$, the five inner layers have dimensions $d^{(1)},\, d^{(2)}, d^{(3)}, d^{(4)}$ and $d^{(4)}$, respectively. The dimension of the outer layer $\hat{X}$ has dimension $d^{(6)}=d^{(0)}$ since this is an auto-encoder. The 5th hidden layer has dimension $d^{(5)}=d^{(1)}$ and the 4th hidden layer has dimension $d^{(4)}=d^{(2)}$. The 3rd layer is the most inner layer with dimension $d^(3)$ which is the reduced dimension we use in some of the frameworks for classification.}
\label{fig: 5layer-auto}
\end{figure}

In Fig.\ref{fig: 5layer-auto}, we show a 5-layer auto-encoder (a neural network with five hidden layers.) We denote the dimension of the $i$th layer by $d^{(i)}.$ The encoder is the composition of the first three inner layers. The third inner layer (the most hidden layer) is the output of the encoder and its dimension is $d^{(3)}$. In two of our experiments, we use this third layer to reduce the dimension of the input data $X$. The input layers $X$ can be either the raw data or the neighbor matrix described in Section~\ref{sec: neigh matrix}.

\section{Classification experiments}\label{sec: experiments}

We include a more granular description on each of the frameworks described in Section \ref{sec: frameworks} that we used on our experiments. 

We have three frameworks consisting of two stages. The first two includes stage 1: perform dimension reduction in raw data with a linear unsupervised method (PCA) or a non-linear unsupervised method (most inner layer of an auto-encoder); stage 2: feed the classifier with new predictors resulting from the dimension reduction. The second two-stage framework includes stage 1: neighbor matrix assembly; stage 2: feed the classifiers with new generated data with features from the neighbor matrix.

The two frameworks with with three stages include stage 1: construction of the neighbor matrix; stage 2: perform dimension reduction in neighbor matrix with linear supervised method (PCA) a non-linear unsupervised method (most inner layer of an auto-encoder); stage 3: feed the classifiers with new predictors resulting from the dimension reduction.

We consider two classifiers, K-nearest neighbors and Random Forest for 6 classes (ground, bridge deck, high noise, water, rail and noise). We choose $k=15$ as the number for nearest neighbors for the construction of the neighbor matrix described in Section~\ref{sec: neigh matrix}. 

We use 100,000 sub-sampled for assembling the neighbor matrix. We chose the latter sub-sample equally spaced according to the order of the original LiDAR data set. We perform two processing steps to the training and testing sets. We basically apply to data transformations: standardization and normalization.

\begin{description}
\item[Step 1.] Standardization of each feature. Compute the mean and standard deviation for the training set and the testing set. Each transformed data set has mean 0 and standard deviation 1. 
\item[Step 2.] Normalization of transformed data sets from Step 1. Re-scaling the training set and testing set to have norm 1. That is, apply the map
\begin{equation*}
x \mapsto \dfrac{x}{{\| x \|}_2},
\end{equation*}

where ${\| \cdot \|}_2$ is the euclidean norm. The maps sends the data points to the points in the unit spheres.
\end{description}

When the dimension reduction stage is inserted in the basic classification framework, we used the explain variance to choose the number of components for PCA and the number of nodes of the inner layer in the auto-encoder. We described the auto-encoder layer terminology in Section~\ref{sec: dim reduct} and, in particular Fig.~\ref{fig: 5layer-auto} to ease understanding. We have two cases depending if we we have the neighbor matrix construction stage:

\begin{enumerate}
	\item If the framework does not include the neighbor matrix construction stage,  we use 5 components for the PCA and a 5 dimensional inner layer of a 5-layer auto-encoder. For the 5-layer auto-encoder, the input layer dimension is $d^{(0)}=7$ (input features: x, y,z,intensity,scan angle,number of returns,number of this return.)  The dimension of the first hidden has dimension $d^{(1)}=6$, the second inner layer has dimension $d^{(2)}=5$ and the most inner layer has also dimension $d^{(3)}=5.$ The dimension layers included in the decoder are $d^{(4)}=5,\, d^{(5)}= 7$ and $d^{(6)}= 7= d^{(0)}.$
	
    \item If the framework includes neighbor matrix (see \eqref{eqn:neigh2} and Fig.\ref{2nd row neigh}), we use 40 components for the PCA and a 40 dimensional inner layer of a 5-layer auto-encoder to perform non-linear dimension reduction. For the 5-layer auto-encoder, the input layer dimension is $d^{(0)}=8(k+1)$, where $k$ is the number of nearest neighbors used to assemble the neighbor matrix. We chose $k=15$ in our experiments. 
     The dimension of the first inner layer is  $d^{(1)}=7(k+1)$, the second layer has dimension $d^{(2)}=5(k+1)$ and the most inner layer has dimension $d^{(3)}=40$ (and the dimension of $E(X)$ where $X$ is the input data.) The dimension layers included in the decoder are $d^{(4)}=5(k+1),\, d^{(5)}= 7(k+1)$ and $d^{(6)}= 8(k+1)= d^{(0)}.$
      In our case, we chose $k=15$ nearest neighbors to generate the neighbor matrix. 
\end{enumerate}

The following parameters used in the auto-encoder implementation. A learning rate of 0.01, 200,000 number of epochs and batch size of 1,000.

In all experiments, the feed forward neural network classifier architecture consists of an input layer made of the new predictors obtained after dimensionality reduction, two hidden layers (first hidden layer has dimension 20, second hidden layer has dimension 15.)

\begin{table}
\begin{tabular}{l |c |c| c | c}
&KNN &RF &RF-Ens &NN \\
\hline 
Raw &0.8670 (+/- 0.0004) &	0.8701 (+/- 0.0007) &	0.8564 (+/- 0.0019) &	0.8241 (+/- 0.0018) \\

PCA &0.8399 (+/- 0.0002) &	0.8384 (+/- 0.0010) &	0.8212 (+/- 0.0011) &	0.7791 (+/- 0.0069) \\

Enc &0.8223 (+/- 0.0004) &	0.8160 (+/- 0.0003) &	0.7902 (+/- 0.0041) &	0.6331 (+/- 0.0110) \\
	

Neig+PCA &0.8291 (+/- 0.0032) &	0.8445 (+/- 0.0029) &	0.8361 (+/- 0.0031) &	0.9748 (+/- 0.0042) \\
	
Neig+Enc &0.7366 (+/- 0.0045) &	0.7816 (+/- 0.0044) &	0.7700 (+/- 0.0049) &	0.6770 (+/- 0.0059) \\
	
Neig &0.8303 (+/- 0.0025) &	0.9497 (+/- 0.0101) &	0.9499 (+/- 0.0118) &	0.9792 (+/- 0.0044) \\	

\hline 

\end{tabular}
\caption{5-fold cross validation of $F_1$ scores for different classification frameworks; number of classes=6; RAW+ Norm= Standardized and normalized  raw data (includes pre-processing step)
Enc= Encoder (using inner layer of auto encoder for dimension reduction);
PCA and Enc have already been standardized and normalized.}
\label{main table}

\end{table}

\begin{table}
\begin{center}	
\begin{tabular}{l | c | c }
           & accuracy	 & error rate\\
\hline            
RF Raw     &  0.8844     &  0.1156 \\
RF Enc     &  0.7637     &  0.2363  \\
KNN PCA    &  0.8416     &  0.1584  \\
KNN Enc    &  0.8391     &  0.1609  \\

NN Neigh   &  0.9847     &  0.0153   \\

NN Neigh + PCA  & 0.9770  &  0.0230  \\
\hline 	

\end{tabular}
\caption{Accuracy and error rates associated to the best f1 scores presented on Table~\ref{main table}}
\label{table: error rates}
\end{center}	
\end{table}

We explain each of the experiments include in Table~\ref{main table}. We are using the following classifiers: K-nearest neighbor (KNN), random forest (RF), and ensemble of random 20 random forests of maximum depth 20, and a feed forward neural network with two hidden layers (NN). The first hidden layer of NN has dimension 20 and he second hidden layer has dimension 15. 

We describe frameworks associated to each row on table Table~\ref{main table} in Table~\ref{table: experiment-description}.

\begin{table}
	
\begin{tabular}{|c|c|l|}
\hline 	
Experiment 1 & ``Raw'' &  The standardized and normalized raw data is directly used as input for each \\
& & of the classifiers mentioned above (KNN, RF, RF-Ens, NN.)\\
\hline 
Experiment 2 &``PCA'' & The input the is the standardized and normalized raw data.\\
& & We first insert the linear dimension reduction stage \\
& & by performing PCA with 5 components.\\
& & We feed each of the classifiers with the new predictors obtained by \\
& & projecting into the subspace generated by the 5 principal components.\\
\hline 	
Experiment 3 & ``Enc'' &  The input the is the standardized and normalized raw data.\\
& & We first insert the non-linear dimension reduction stage by using\\
& & the most inner layer (the third one) of the 5-layer auto-encoder. \\
& & The dimension of the most inner layer is $d^(3)=5$.\\
& & We feed each of the classifiers with the new predictors obtained by projecting into\\
& & the manifold generated by the encoder, $E(X).$\\
\hline 
Experiment 4 &  ``Neigh + PCA'' & The input the is the standardized and normalized neighbor matrix \\
& & (assembled with 100,000 examples.) \\
& & We first insert the linear dimension reduction stage by performing\\
& & PCA with 40 components.\\
& & We feed each of the classifiers with the new predictors obtained by projecting into\\
& & the subspace generated by the 5 principal components.\\
\hline 	
Experiment 5 &  ``Neigh + Enc'' & The input the is the standardized and normalized neighbor matrix \\
& & (assembled with 100,000 examples.)\\
& & We first insert the non-linear dimension reduction stage by using\\
& & the most inner layer (the third one) of the 5-layer auto-encoder. \\
& & The dimension of the most inner layer is $d^(3)=40$.\\
& & We feed each of the classifiers with the new predictors obtained by projecting into\\
& &  the manifold generated by the encoder, $E(X).$\\
\hline 
Experiment 6 &  ``Neigh'' & The standardized and normalized neighbor matrix \\
& & (assembled with 100,000 examples) is directly used as\\
& & input for the classifiers mentioned above (KNN, RF, RF-Ens, NN.)\\
\hline 		
\end{tabular}
\caption{Description of experiment. The cross-validated $F_1$ scores for these experiments are presented in Table~\ref{main table}.}
\label{table: experiment-description}
\end{table}

We defined the $F_1$ metric in \ref{f1-score}. Table~\ref{main table} shows the 5-fold cross validated scores as described in Section~\ref{sec: frameworks}.

In Table~\ref{main table},the highest 5-CV-$F_1$ scores are observed when using the neighbor matrix with random forest, the ensemble of random forest and the neural network. Also, we also observe a high scores (0.9748) when combining a neighbor matrix (previously standardized and normalized), performing PCA and then using the feed forward neural network classifier.

We also note that using the neighbor matrix as input and using the inner layer of the auto-encoder does not perform as well as the combination neighbor matrix and auto-encoder. On the other hand, observe that for classifiers KNN and RF, using raw data as input and then reducing the dimension with the encoder gives similar results as when using neighbor matrix as input and reducing the dimension with PCA.

Table~\ref{table: error rates} includes the accuracy and error rates for the best $F_1$ scores as observed in Table~\ref{main table}. Notices that the error rate corresponding to the neighbor matrix with and without inserting PCA on the framework is at least six times less than the error rate corresponding to the rest of the methods. 

\begin{figure}
\begin{center}	
\includegraphics[scale=0.5]{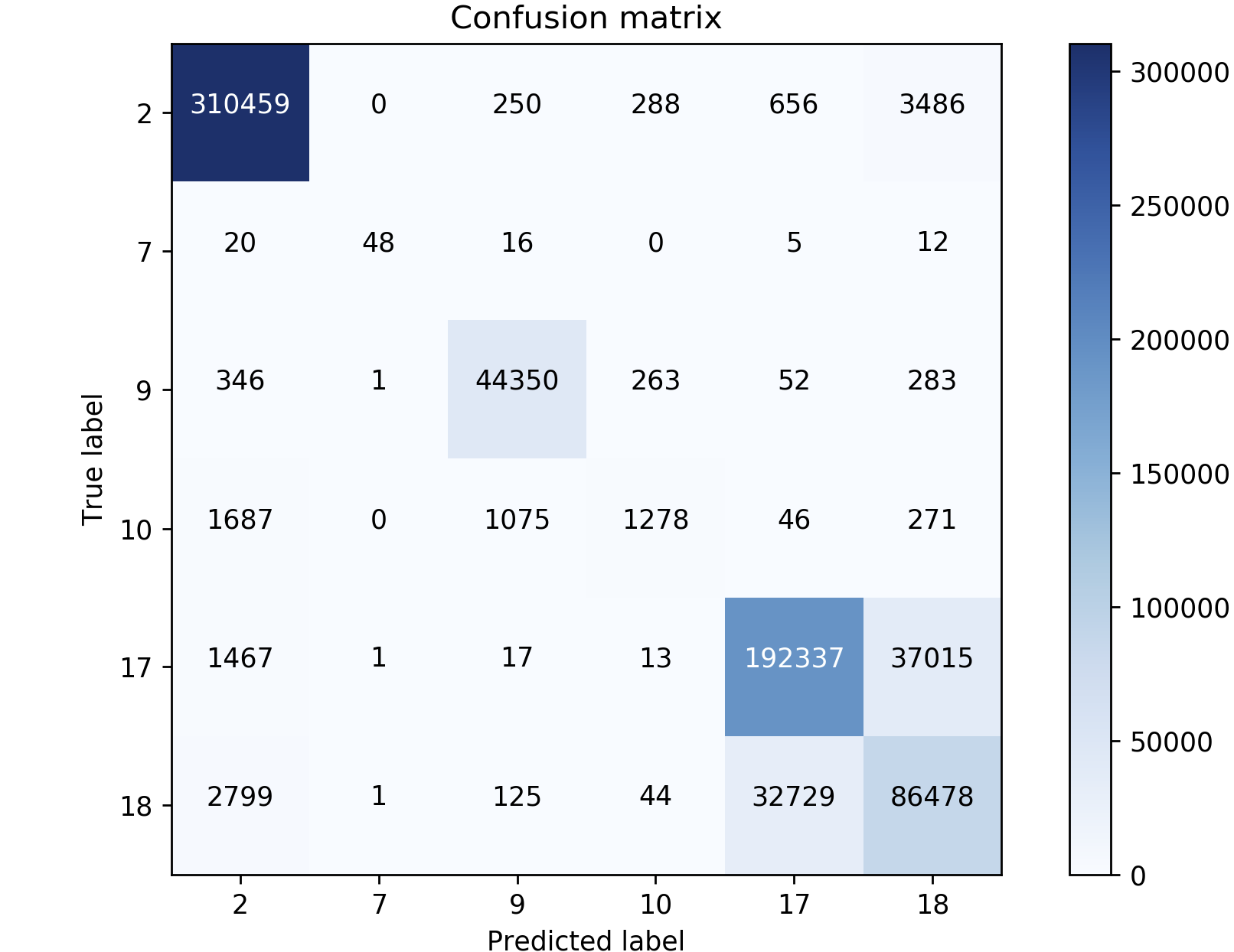}
\end{center}
\caption{Confusion matrix corresponding to the random forest classifier with raw data as input}
\label{fig: matrix-raw-rf}
\end{figure}

\begin{figure}
	\begin{center}	
		\includegraphics[scale=0.8]{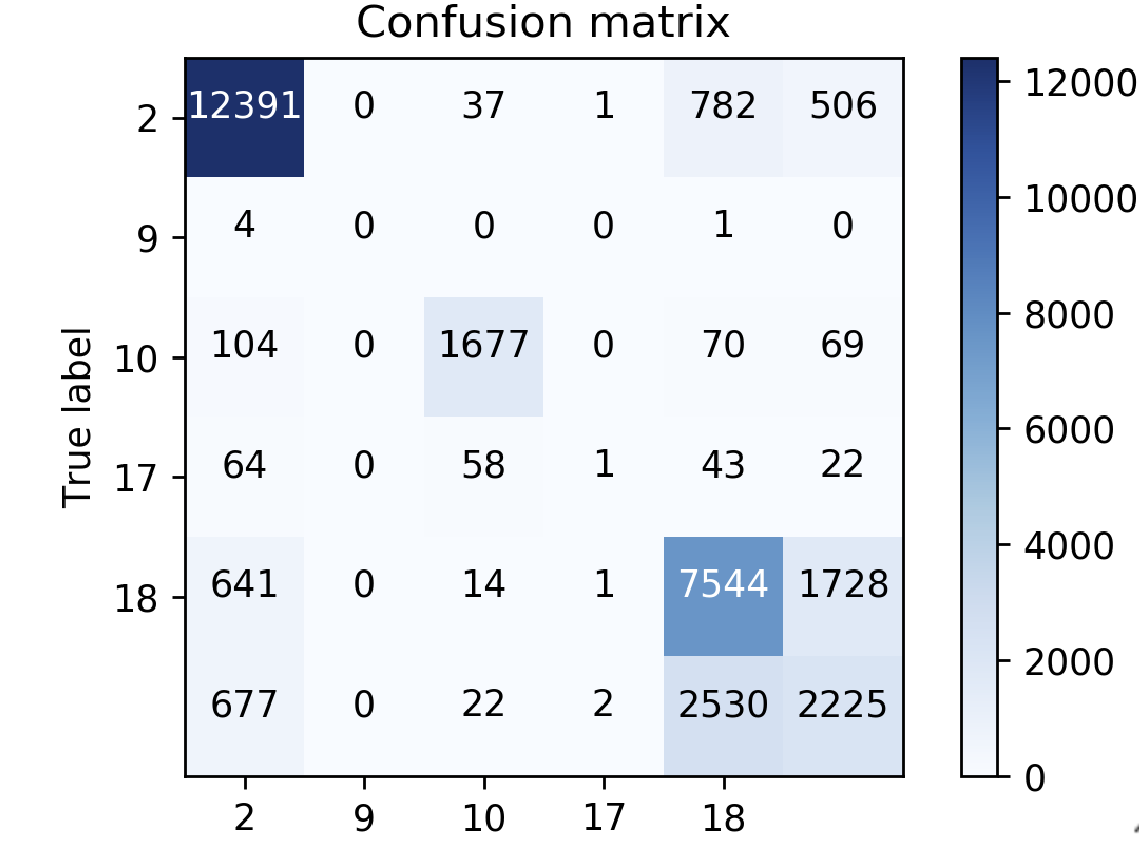}
	\end{center}
	\caption{Confusion matrix corresponding to the random forest classifier with new predictors originated from the inner layer of the auto-encoder as input}
	\label{fig: matrix-enc-rf}
\end{figure}

\begin{figure}
	\begin{center}	
		\includegraphics[scale=0.5]{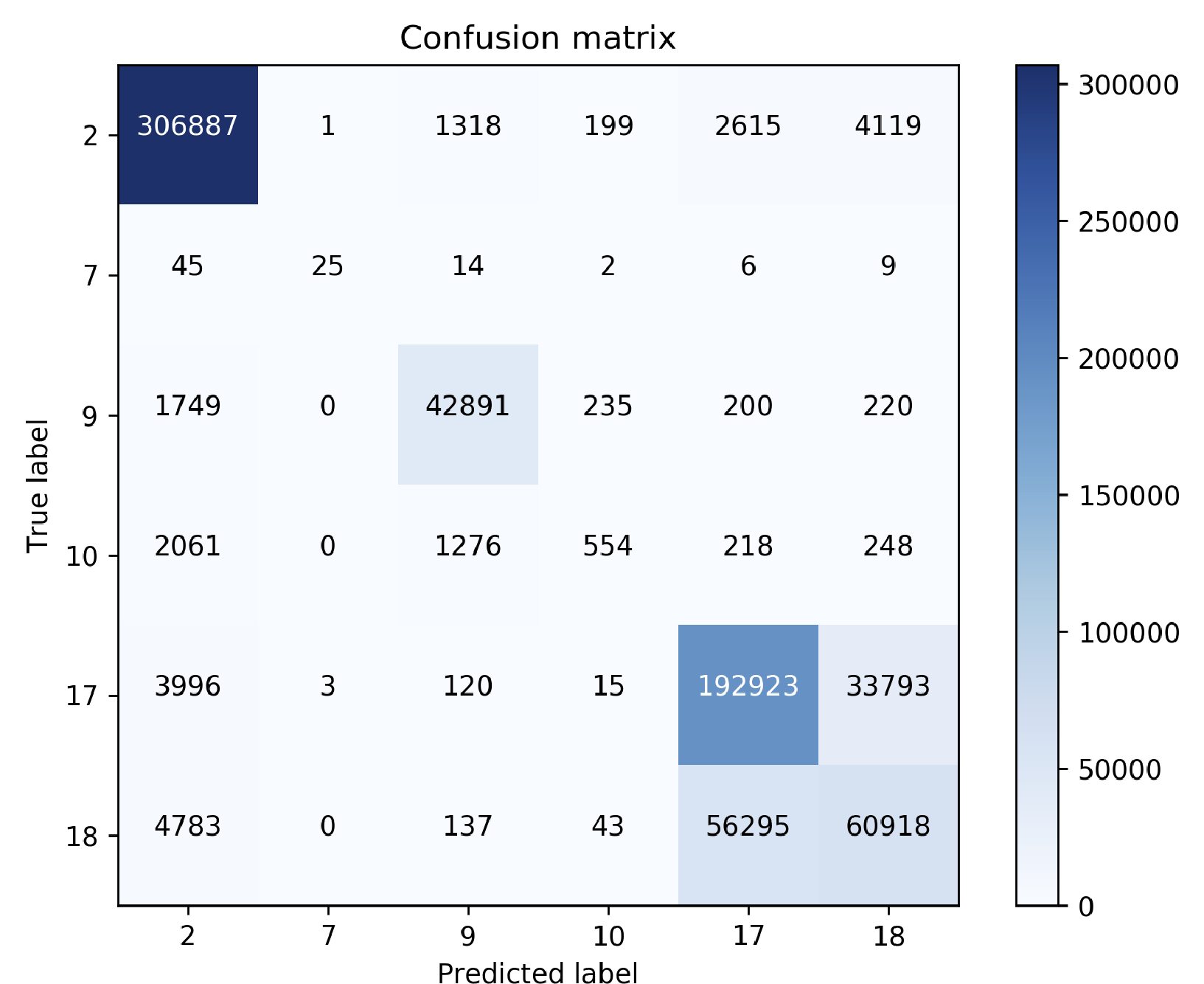}
	\end{center}
	\caption{Confusion matrix corresponding to k-nearest neighbor classifier with predictors originated from PCA as input}
	\label{fig: matrix-pca-knn}
\end{figure}

\begin{figure}
	\begin{center}	
		\includegraphics[scale=0.5]{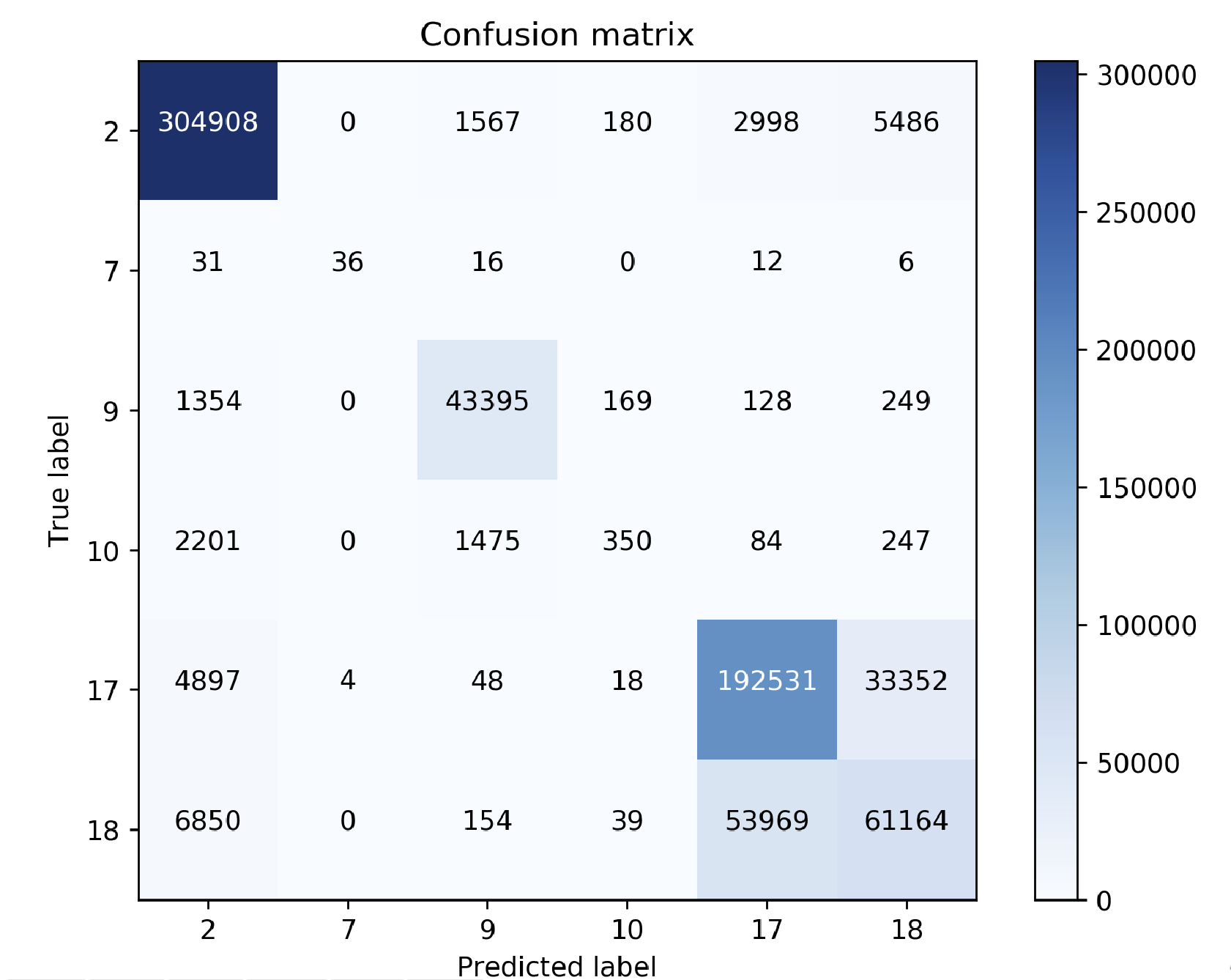}
	\end{center}
	\caption{Confusion matrix corresponding to k-nearest neighbor classifier with predictors originated from the inner layer of the auto-encoder as input}
	\label{fig: matrix-enc-knn}
\end{figure}

\begin{figure}
	\begin{center}	
		\includegraphics[scale=0.5]{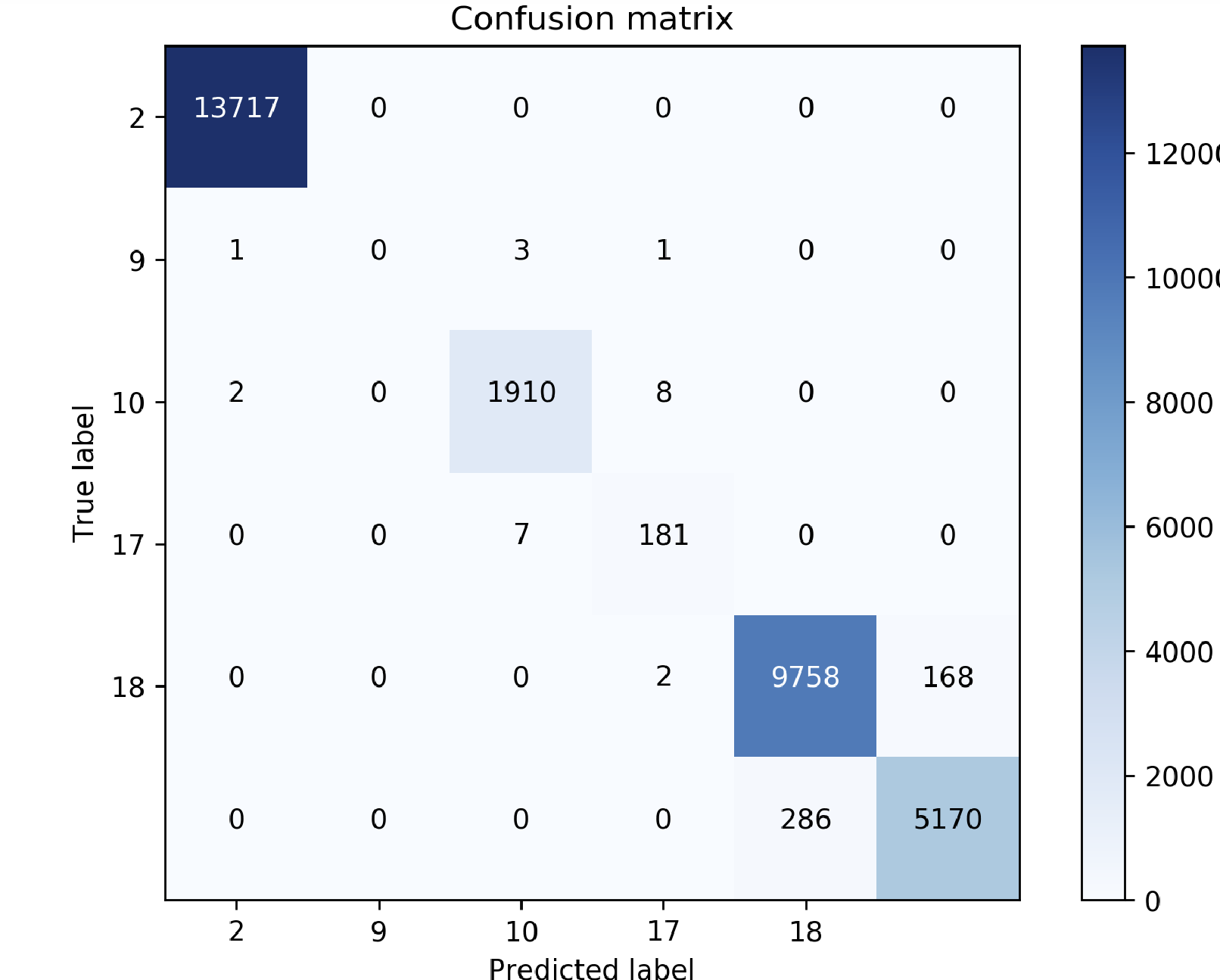}
	\end{center}
	\caption{Confusion matrix corresponding to feed-forward neural network classifier with the neighborhood matrix as input}
	\label{fig: matrix-Neigh-NN}
\end{figure}

\begin{figure}
	\begin{center}	
		\includegraphics[scale=0.5]{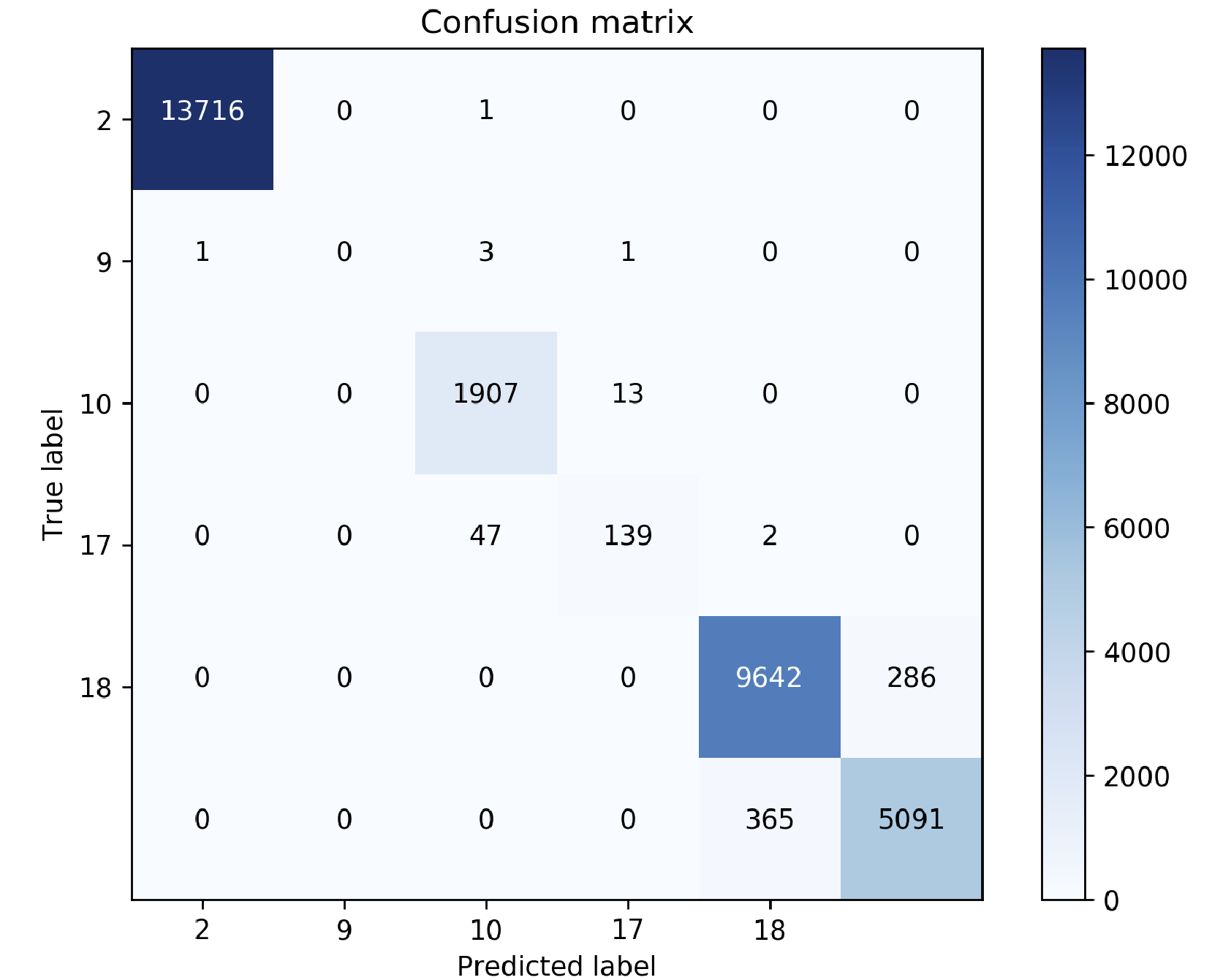}
	\end{center}
	\caption{Confusion matrix corresponding to feed-forward neural network classifier with the new predictors originated from PCA applied to the neighbor matrix as input}
	\label{fig: matrix-Neigh-PCA-NN}
\end{figure}

We included the confusion matrices corresponding to the highest f1-score for each case on Figures~\ref{fig: matrix-raw-rf}--\ref{fig: matrix-Neigh-PCA-NN}.

\section{Summary and Future Research Directions}
\label{summary}

We performed a comparison of various classification techniques using linear dimension reduction (PCA) and non-linear dimension reduction (auto-encoder.) The best results ($F_1$ scores) were obtained by using the neighbor matrix as input and the reducing the dimension of the new data frame using PCA and using a feed forward neural network as classifier. Moreover, using a feed forward neural networks as classifier applied to the neighbor matrix with and without inserting the PCA step shows great improvement in the error rates respect to the other frameworks. Improving the performance of a classifying framework to differentiate elementary classes such as vegetation, water, ground, etc. will help to automate processes on applications such habitat mapping, elevation models among others.

The research effort revealed a number of potential future research directions: 

\begin{itemize}
\item  Exploiting intrinsic dimension techniques at different scales to generate more features. In this way, the algorithm will have more information on the geometry of the data to perform better classification of the classes. See \cite{Fukunaga198215ID} for work in estimation of intrinsic dimension using local PCA and \cite{BRODU} for a multi-scale classification example using support vector machines. \cite{LevinaB04} and \cite{lebi2005} provides a maximum likelihood framework for intrinsic dimension estimation. 

\item Determine relationships between encoder-decoders and product coefficient  representations of measures

\item Analyze a larger forestry data with trees and classes such as trunk, ground and leaves. This is linked to an important application related to climate change. See \cite{Moskal} for definition and theories of indirect and direct methods to estimate the leave to area (LAI) index in terrestrial LiDAR which is relevant to the gas-vegetation exchange phenomenon understanding. 

\item Modify the architecture of the auto-encoder by adding more layers and/or changing the dimension of the inner layers. Compare the accuracy using this new preprocessing step with the one resulting from PCA.

\item Perform shape analysis by combining the results from this paper with the current shape analysis state-of-the art techniques. The application would use the shape recognition in forestry data where the recognition of leaf shapes would be of great interest for practitioner.

\end{itemize}

\begin{acknowledgement}
This research is supported by the Azure Microsoft AI for Earth grant. Many thanks to Monika Moskal (WU) Jonathan Batchelor (WU) and Zheng Guang (NU) for sharing their expertise in the technical aspects of LiDAR data acquisition and for encouraging pursuing the next future directions for application in forestry.

We gratefully acknowledge Linda Ness for encouraging further discussions on manifold learning for LiDAR data in the \emph{Women in Data Science and Mathematics Research Collaboration Workshop (WiSDM)}, July 17-21, 2017, at the \emph{Institute for Computational and Experimental Research in Mathematics (ICERM)}. The workshop was partially supported by grant number NSF-HRD 1500481-AWM ADVANCE and co-sponsored by Brown's Data Science Initiative.

\end{acknowledgement}

\bibliographystyle{siam} 
\bibliography{references_lidar_auto}

\end{document}